\documentclass{article}

\usepackage[utf8]{inputenc}
\usepackage{authblk}

\title{Catastrophic Importance of \\ Catastrophic Forgetting}
\author{Albert Ierusalem \\ albertierusalem@protonmail.com}
\affil{Independent Researcher}
\date{}

\usepackage{natbib}
\usepackage{graphicx}
\usepackage{amsmath}
\usepackage{algorithm,algorithmic}
\usepackage{multicol}
\usepackage{titlesec}
\usepackage{blindtext}
\usepackage{color}
\usepackage[dvipsnames]{xcolor}
\usepackage{amsfonts}
\usepackage{parskip}
\usepackage{fancyhdr}
\graphicspath{ {images/} }
\usepackage{etoolbox}

\begin{document}

\maketitle

%\begin{multicols}{2}
\begin{abstract}
This paper describes some of the possibilities of artificial neural networks that open up after solving the problem of catastrophic forgetting. A simple model and reinforcement learning applications of existing  methods are also proposed

\end{abstract}

\section{Introduction}
There are a number of forms of forgetting in the human brain, and this is a normal, adaptive, and necessary process for learning. One of the more interesting is the active form that helps us concentrate on some special task by moving all information that is unnecessary at that moment to unconscious memory. Hermann Ebbinghaus [2], a German psychologist, became the first person to study memory experimentally. In his early experiments, he encountered the problem of proving that something has been completely forgotten because, allegedly, forgotten information can still affect behaviour and is quite often remembered later. As a result, a definition was proposed that forgetting is the inability to extract from memory at a particular given moment something that was readily extracted from memory earlier.

Later experiments on memory were conducted on the mollusc Lymnaea stagnalis [15] due to the fact that their nerve cells are rather large, with many of those nerves identified and their functions described. There is one nerve cell without which L. stagnalis cannot learn a new skill. If this cell is destroyed, the molluscs not only lose the ability to learn, they also do not forget previously learned behaviours. The model in this study proposes to translate this function into artificial neural networks and define the architecture with active forgetting mechanisms, which gives the name of the model, active forgetting machine (AFM). The AFM contains special neural networks that allow temporary forgetting of unnecessary information by disabling unwanted neurons, and then other combinations of neurons are activated to learn and solve some task.

In the classical interpretation of artificial neural networks, all neurons in the hidden layer are initially activated, and in order to concentrate on a specific task, it is necessary to turn some of them off; in other words, it is necessary to ‘forget’ all unnecessary information. In the context of artificial neural networks, activation means that the neurons are involved in forward propagation during evaluation and backward propagation during training.

\section{Related work}

One of the most interested method that overcoming catastrophic forgetting in neural networks, is EWC [7], where learning of this model consists of adjusting the sets of weights and biases of the linear projections to optimize performance. PathNet method [3] preassigns a level of network capacity per task, and progressive neural networks (PNNs) [14] distribute the network weights in a column-wise fashion, preassigning a column width per task. Similarly, PackNet [8] employs a binary mask to constrain the network. The HAT model [16] also focuses on network weights, but to constrain them, it uses unit-based masks, which also results in a lightweight structure. It avoids any absolute or preassigned pruning ratio, although it uses the compressibility parameter c to influence the compactness of learned models. Context-dependent gating [9], another neuroscience-inspired solution, can further support continual learning when combined with synaptic stabilization.

This paper considers a general class of models designed to solve the problems of catastrophic forgetting. Most of all exiting methods can be generalized to active forgetting mechanism, an important change is that the mechanisms of active forgetting can independently activate the necessary neurons for a specific task.

\section{Active Forgetting Machines}

\subsection{Notation}
While multitasking ability allows the proposed model to switch between several problems, it is also useful during the solving of a single problem. Almost any task can be hierarchically divided into sub-tasks, and the depth of such partition, increases with the complexity of the basic task. Achieving a goal in such multilevel environments is a problem. When some mechanisms of active forgetting are introduced, model can simplify goal achieving by breaking tasks into simpler steps and training a separate combination of neurons for each sub-task. This trick naturally increases the ability of the model to select the correct action.

To describe a general class of active forgetting mechanisms, purposed Active Forgetting Machine model is composed of forgetting net $V$, associative controller $C$, and forgetting algorithm $E$, used to find the best minimal combination of necessary neurons $F$. With this model, $C$ is trained to activate the correct neurons $F$, allowing $V$ to concentrate on a specific task $M$. Forgetting net $V$ uses forgetting layers that are capable of applying multiplied layers of neurons on a binary mask both forward and backward.

Associative controller $C$ is a neural network with an output layer size having the same number of neurons as $V$ forgetting layers, where $C$ is trained to emit mask $F_t$, defined by  algorithm $E$ whenever it receives any sample of task $M$.

For example, if the problem to be solved involves two coordinately different tasks $A$ and $B$, the sequence of actions that will lead to each goal are also different. To solve the two problems, the model should clearly define which groups of neurons, $F_a$, are responsible for the performance of actions in task $A$. Having determined this, the model trains a completely different group of neurons, $F_b$, to achieve the goal in task $B$. Once the groups of neurons are defined, depending on the situation, the model can switch between the strategies, activating different groups of neurons. The learning algorithm for the AFM is shown in Algorithm 1.

\begin{algorithm*}[t]
    \caption{AFM}
    \label{AFM_training}
    \begin{algorithmic}[1]
        \REQUIRE Initialize $V_{\theta}, C_{\phi}$ with random weights $\theta, \phi$; Initialize algorithm $E$;
        \\Train $V_{\theta}$ for first task $T_{0}$ with fully connected mask;
        \medskip
        \STATE By $E$ choose best set $F_{0}$ of activated neurons using pre-trained $V_{\theta}$ for $T_{0}$;
        \STATE Train network $V_{\phi}$ output set $T_{0}$ for task $T_{0}$;
        \REPEAT
        \FOR{$t$ in tasks $T$}
          %\STATE Choose best set of activated neurons $F_{t}$ by $E$ using $V_{\theta}$ and $t$; 
          \FOR{$e$ in epochs}
            %\FOR{$t$ in tasks $T$}
            \STATE Train network $V_{\theta}$ for task $t$, with set $F_{t}$;
            \STATE Train network $C_{\phi}$ output set $F_{t}$ for task $t$;
            %\ENDFOR
          \ENDFOR
        \ENDFOR \UNTIL{convergence}
    \end{algorithmic}
\end{algorithm*}

\subsection{Not Bayesian Variational multitask learning}

\begin{figure*}[t]
  \centering\includegraphics[scale=0.60]{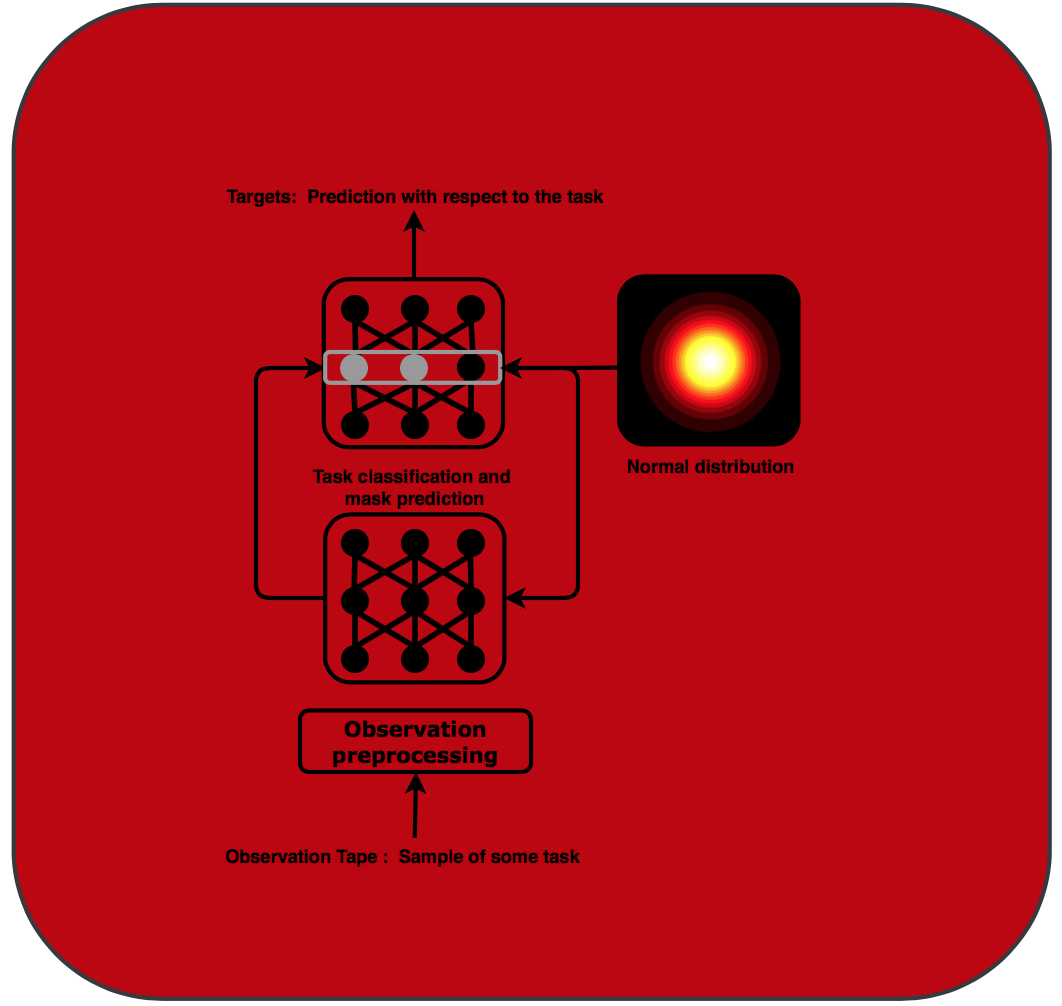}
  \caption{AFM Architecture with Variational Dropout mask selection}
\end{figure*}%

One of the way to find the best combination of activated neurons, $F_t$ in some task $t$, it is to make neural network connections sparse, and estimate which architecture is the most useful.

In Bayesian Learning we usually have some initial belief over parameter $w$, in the form of a prior distribution $p(w)$. After observing data $D$, using Bayesian Inference, prior distribution can be transformed into a posterior distribution $p(w | D) = p(D | w)p(w)/p(D)$. Is necessary to use approximation techniques, because computing the true posterior distribution using the Bayes rule usually involves computation of intractable integrals.

Approximation of the posterior distribution $p(w | D)$ by optimize parameters $\phi$ of some parameterized model $q_\phi(w)$ is called variational inference. Measure of such approximation is the Kullback-Leibler divergence $D_{KL}(q_\phi(w)|p(w | D))$. In practice, optimal value of variational parameters $\phi$ can be found by maximization variational lower bound $L(\phi)$ of data marginal likelihood:
\begin{equation}
    \mathcal{L}(\phi)= -D_{KL}(q_\phi(w)|p(w)) + L_{D}(\phi)
\end{equation}
\begin{equation}
    L_{D}(\phi) =\sum_{(x,y)\in D} \mathbb{E}_{q_{\phi}}[log(y|x,w)]
\end{equation}

The trick to estimate variational lower bound, and compute it gradients was presented by Kingma and Welling [6]. The main idea is to represent the random parameters $q_{\phi}(w)$ as a deterministic differentiable function $w = f (\phi, \varepsilon )$ where $f(\cdot )$ is a differentiable function and $\varepsilon  \sim  p(\varepsilon )$ is a non-parametric noise. This parameterisation make  possible to obtain an unbiased differentiable minibatch-based Monte Carlo estimator of the expected log-likelihood. This method is called stochastic gradient variational Bayes
\begin{equation}
    L_{D}(\phi) \simeq L_{D}^{SGVB}(\phi) = \frac{N}{M}\sum_{(i=1)}^{M} log(y^i|x^i, w = f (\phi, \varepsilon ))
\end{equation}
where $\varepsilon$ is a noise vector drawn from the noise distribution $p(\varepsilon)$, and $(y^i|x^i)^{M}_{i=1}$ is a minibatch of data $D$. To reduces the variance of this gradient estimator, was presented Local Reparametrization trick [1]. The idea is to sample noise to activations for each data-point inside mini-batch, and this method was noted as not Bayesian [4].

But the Local Reparametrization trick and Variational dropout techniques has been used to obtain sparse networks [11]. in Variational Dropout, $q(W | \theta, \alpha)$ is using as an approximate posterior distribution for a model. Where weights $w_{ij}$ is random variable parametrized by $\theta_{ij}$, 
\begin{equation}
    w_{ij}=\theta_{ij}\varepsilon_{ij}=\theta_{ij}(1+\sqrt{\alpha_{ij}} \cdot \epsilon_{ij})\sim \mathcal{N}(w_{ij}|\theta_{ij},\alpha\theta_{ij}^2), \epsilon_{ij} \sim\mathcal{N}(0,1). 
\end{equation}
It is difficult to train network with Variational Dropout, because of a large variance of stochastic gradients when $\alpha \geq  1$.
\begin{equation}
    w_{ij}=\theta_{ij}(1+\sqrt{\alpha_{ij}} \cdot \epsilon_{ij}), \frac{\partial w_{ij}}{\partial \theta_{ij}} = 1+\sqrt{\alpha_{ij}} \cdot \epsilon_{ij}, \epsilon_{ij} \sim\mathcal{N}(0,1). 
\end{equation}

To avoid it, was proposed a trick [11]  which replacing multiplicative noise term $1+\sqrt{\alpha_{ij}} \cdot \varepsilon_{ij}$ with an exactly equivalent additive noise term $\sigma_{ij} \cdot \varepsilon_{ij}$ , where $\alpha_{ij}^2 = \alpha_{i,j}\theta{ij}^2$. 
\begin{equation}
    w_{ij}=\theta_{ij}(1+\sqrt{\alpha_{ij}} \cdot \epsilon_{ij})=\theta_{ij}+\sigma_{ij}\cdot\epsilon_{ij},  \frac{\partial w_{ij}}{\partial \theta_{ij}} = 1,   \epsilon_{ij} \sim\mathcal{N}(0,1)
\end{equation}
Using this trick, model can be trained within the full range of $\alpha_{ij}$. For KL divergence that is tight for all values of $\alpha$ was proposed approximation, with $k_1=0.63576, k_2=1.87320, k_3=1.48695$

\begin{equation}
    -D_{KL}(q_\phi(w_{ij}|\theta_{ij},\alpha_{ij})|p(w_{ij})) \approx k_1\sigma(k_2+k_3log\alpha_{ij})-0.5log(1+\alpha^{-1}_{ij}) + C
\end{equation}

During this, the posterior over this weight is a high-variance normal distribution, if $\alpha_{ij}\rightarrow \infty$ for a weight $w_{ij}$. It is beneficial for model to put $\theta_{ij} = 0$ as well as $\sigma{ij} = \alpha_{ij}\theta^{2}_{ij} = 0$ to avoid inaccurate predictions. As a result, the weight which does not affect the network’s output can be ignored, because the posterior over $w_{ij}$ approaches zero-centered $\delta$-function. 

For the AFM model, the mask on the neural layer is formed depending on the number of activated bonds for a specific neuron. For each neuron, if the number of input weights $w_{ij}$ with $\alpha \geq  1$, greater than threshold drop hyperparameter, which was set equal to $97\%$, such a neuron is ejected and labeled with zero. %Experiments with such method are demonstrated in next section.

\subsection{Supervised experiment}

In this experiment, the goal of the model was to learn, in a supervised manner, to classify an Fashion digits MNIST data sets, where the input was 28 by 28 pixels in a 1-d vector for fully connected controller and actor networks. The output of the model was ten classes, the same number in each data set. The main purpose of this classification problem was to show the ability of the AFM to maintain high accuracy while working on various tasks. The results are shown in Table 1.

The results were compared with the HAT model, which showed better results on multitasks than PathNet, PNN, EWC, and other models for overcoming catastrophic forgetting. Due to sparse connections, AFM is not sensitive to changes in architecture. After training, the AFM model had 62, 93 and 116 activated neurons in hidden layer. Also, an important difference is that the AFM has common output layer for both tasks.   

\begin{table}
\centering
\begin{tabular}{llll}
\hline
Model & Architecture &MNIST Accuracy & Fashion MNIST accuracy \\ \hline
      
AFM    &(784, 128, 10)              & 0.95           & 0.87                    \\ 
HAT    &(784, 128, 128,  (10,10))   & 0.91           & 0.83                     \\ 
EWC    &(784, 128, 128,  (10,10))        & 0.97           & 0.79                     \\
SGD    &(784, 128, 128, (10,10))        & 0.50           & 0.87                   \\ \hline
AFM    &(784, 800,  10)             & 0.96           & 0.86                    \\ 
HAT    &(784, 800, 800,  (10,10))   & 0.97           & 0.87                     \\ 
EWC    &(784, 800, 800,  (10,10))   & 0.83           & 0.90                     \\
SGD    &(784, 800, 800,  (10,10))   & 0.44           & 0.90                    \\ \hline
AFM    &(784, 2000, 10)             & 0.95           & 0.85                   \\ 
HAT    &(784, 2000, 2000,  (10,10)) & 0.98           & 0.90                     \\ 
EWC    &(784, 2000, 2000,  (10,10)) & 0.88           & 0.91                     \\
SGD    &(784, 2000, 2000,  (10,10)) & 0.71           & 0.91                    \\ \hline
%PathNet   & \textbf{0.99}           & \textbf{0.91}                     \\ \hline

\end{tabular}
\caption{Models accuracy for MNIST and Fashion MNIST at same time.(in progress)} %Where $E_{E}$ is as Evolutionary algorithm for best mask search and $E_{V}$ is Variational mask method.}
\end{table}

\section{Reinforcement Learning applications}

\subsection{Forgettable Decision Processes}
The Markov Decision Processes MDP formally describe an environment for reinforcement learning, where the problem is to find policy $\pi$ to reach the goal. A MDP is a tuple $\left \langle S,A,P,R,\gamma \right \rangle$ where:
\begin{itemize}
  \item $S$ is a set of states
  \item $A$ is a set of actions
  \item $P$ is a state transition probability matrix, \\ $P_{ss^{'}}^{a}=\mathbb{P}[S_{t+1}=s^{'} \mid S_{t}=s, A_{t}=a]$
  \item $R$ is a reward function, \\
  $R_{s}^{a}=\mathbb{E}\left [ R_{t+1} \mid S_{t}=s, A_{t}=a \right ]$
  \item $\gamma$ is a discount factor $\gamma \in \left [ 0,1 \right ]$
\end{itemize}

A particular MDP is defined by its state and action sets and by the one-step dynamics of the environment. Most environments have a different nature of reward, if take it into account, the policy $\pi$ should be different for each reward too. Based on the paradigm of active forgetting, presented Forgettable Decision Processes framework which is adapted to a variety of policies.
The FDP is a 4-tuple $\left \langle M,F,P,R, \right \rangle$ where:
\begin{itemize}
  \item $M$ is a set of MDP's for each of the task 

  \item $F$ is a set of structural-functional units which available to be used in MDP $M_{t}$, where $t$ is a number of task.

  \item $P$ is a MDP to MDP transitions probability matrix
  \item $R$ is a rewards received during $M_t$ to $M_{t+1}$ transition

\end{itemize}
In the context of artificial neural networks, each $F_{t}$ can be presented as a set of neurons.Each set of structural-functional units $F_{t}$ and MDP $M_{t}$, allows to perform a certain sequence of actions. And set of states $S$ can also be described by MDP $M_{t}$.  
So $P$ can be presented as: 
\begin{equation}
  P_{mm^{'}}^{f}=\mathbb{P}[M_{t+1}=m^{'} \mid M_{t}=m, F_{t}=f ] 
\end{equation}
And reward $R$, where $R_{m}^{f}$ is set of all rewards which was get in MDP $M_{t}$ and mask $F_{t}$ 
\begin{equation}
  R_{m}^{f}=\mathbb{E} [ R_{t} \mid M_{t}=m, F_{t}=f, ]
\end{equation}
If we consider the environment with a variety of rewards, as divided into subtasks, we can move from space of actions to space of $F$ structural-functional units, and train set of neuron $F_{t}$ to specific MDP $M_{t}$, to switch between policies $\pi_{t}$.

\subsection{Evolutionary multitask learning}
The evolutionary algorithm $E$ is necessary to find the most productive and unique combination of neurons $F_t$ in $V$ for the exact task $M_t$. $E$ is a simple evolutionary algorithm which consists of $initialization$, $selection$, $crossover$, $mutation$ and $termination$, where selection is carried out by applying $F_t$ in the forgetting network $V$.

First, the $V$ network is trained to perform some task $M_t$, and when a threshold accuracy value is received, the evolutionary algorithm $E$ begins searching the largest groups of neurons that are least effectively accurate at the task and enables retention of intact neurons for another task. The most unnecessary neurons are marked zero, and the result is a mask vector $F_t$ of ones representing important neurons. When the mask $F_t$ is completed, it is time for $C$ network.

For each new task, the entire training cycle is repeated, taking into account that omitted masks should be minimally crossed with the previous ones.

\subsection{Algorithmic experiment}
The algorithm consisted of two sub-tasks, $Copy$ and $Drop duplicate$, which differ in the root. The input of the agent consists of the state, which is represented by the current input symbol, the last step’s input symbol, and the last step’s reward. The size of the agent output layer is the size of the vector of possible symbols. In the first sub-task, $Copy$, the goal of the model is to repeat some sequence, and if the last reward and last actions of this sub-task are labelled as zero vectors, the information is unnecessary for the agent in this sub-task, which helps controller $C$ find differences between sub-tasks. The input in this sub-task was the sequence ADEBCADEBCEEDBACBAEBBDCAECBACBEAEAEBC, and the goal of the model was a symbol-by-symbol repeat of the input sequence.

In the second sub-task, $Dropduplicate$, the goal of the model was to repeat the input, dropping duplicates. For example, if the input sequence is AAADDDEEEEBBBCCAAAADDEEDDBBBBAACCBBAAEEBBBDDCC, the target of the model is ADEBCADEDBACBAEBC. This sub-task input consists of the current state, last input, and last reward.

\begin{figure*}[t]
  \centering\includegraphics[scale=0.60]{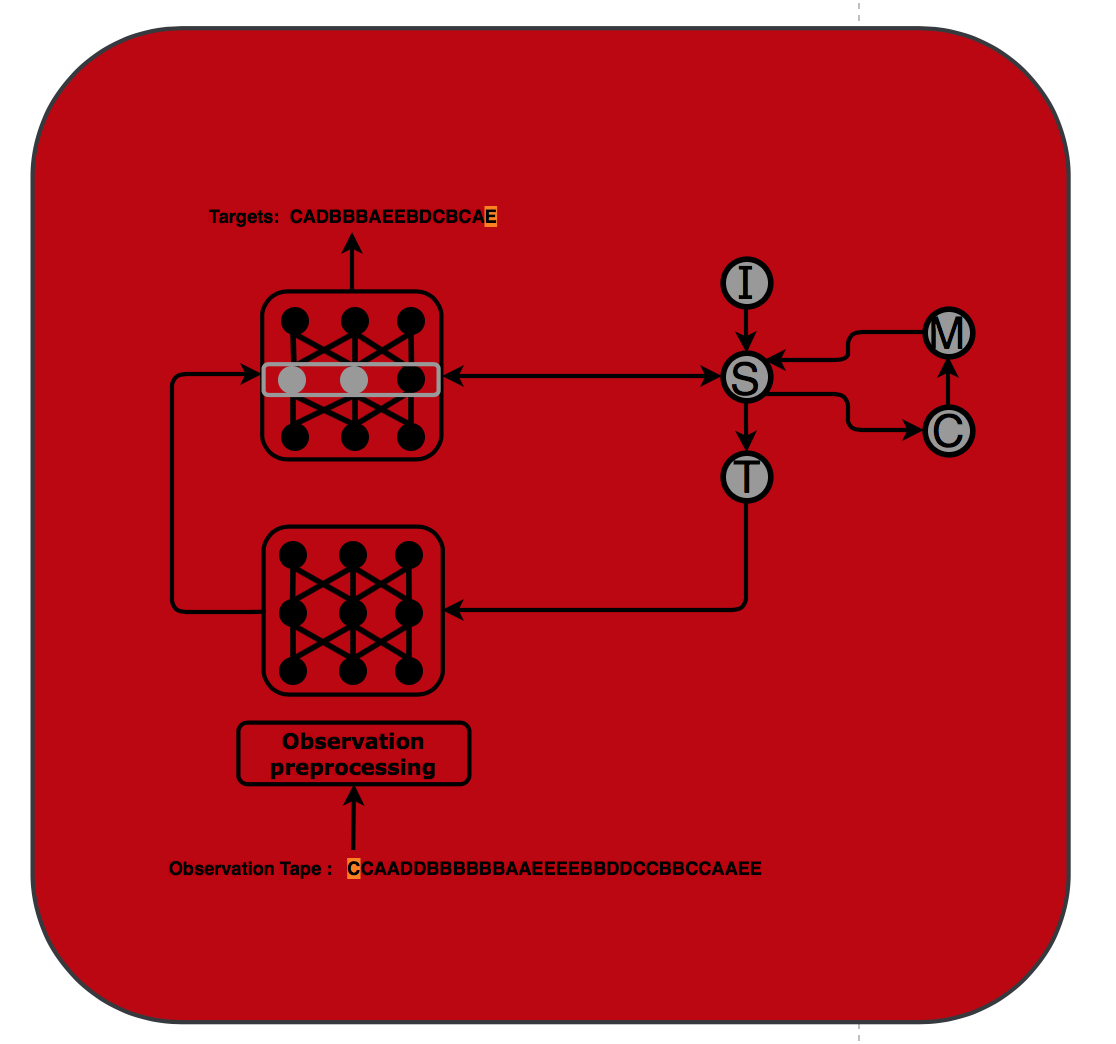}
  \caption{AFM Architecture for Algorithmic task with evolutionary mask. At the bottom demonstrated the controller network, which based on input, choose the activated neurons in forgetting network. At the right demonstrated the evolutionary algorithm, which construct best set of activated neurons}
\end{figure*}%

The model is trained for the first sub-task, $Copy$, and when an average of 20 rewards over 100 trials is reached, the evolutionary algorithm $E$ starts searching for the best combination of activated neurons, performing selection by received rewards $R_{m}^{f}$. Then, this combination, $F_t$, and the opposite combination are used as targets for controller $C$. When the model has masks for each sub-task, the training with fixed neurons begins for the $Copy$ sub-task. At this point, the model can solve tasks independently because of the deep Q learning [10] used for training the model with insight from each of the MDPs. Combining MDP and FDP, to describe the learning process, the optimal action-value function will look like:

\begin{equation}
    Q^{*}\left (m, f, s, a \right ) = max_{\pi_{m}} \mathbb{E}\left [ R_{t} |M_{t}=m, F_{t}=f, S_{t}=s, A_{t}= a,\pi_{m}  \right ]
\end{equation}

$\pi_{m}$ is a policy mapped to actions to MDP, $M_{t}$.
To built optimal strategy, which select the actions to maximize the expected value $r+\gamma max_{a^{'}}Q^{*} (m, f, s^{'},a^{'} )$, is using $Bellman$ $Equation$ with respect to $m$ and $f$
\begin{equation}
    Q^{*} (m, f, s, a  ) = \mathbb{E}_{s^{'}\sim \varepsilon } [ r+\gamma max_{a^{'}}Q^{*} (m, f, s^{'},a^{'}  ) |m,f,s,a ] 
\end{equation}
The neural network is using like a non-linear approximation to action-value function,
\begin{equation}
    L_{i}(\theta_{i}^{f})=\mathbb{E}_{s^{'}, a\sim  \rho  ( \cdot   ) } [  ( y_{i} -Q   (m, s, a; \theta_{i}^{f}  )  )^{2} ]
\end{equation}
Where $\theta_{i}^{f}$ is constructed by mask $f$.
Finally the loss function for the $\theta$ parameters update is:
\begin{align}\label{eq:6}
    \begin{split}
     \bigtriangledown_{\theta_{i}^{f}} L_{i}(\theta_{i}^{f}) &= \mathbb{E}_{s^{'}, a\sim  \rho ( \cdot   ); s^{'}\sim \varepsilon  } [  ( r+\gamma max_{a^{'}}Q^{*} (m, s^{'},a^{'} ; \theta_{i-1}^{f} )\\
     &- Q(m, s,a;\theta_{i}^{f} )) \bigtriangledown_{\theta_{i}^{f}} Q (m, s,a;\theta_{i}^{f}  )  ]
    \end{split}
\end{align} 
During training for $Drop duplicate$ subtask, neurons for $Copy$ subtask was fixed with mask, emitted by associative controller $C$. 

\begin{figure*}[t]
  \centering\includegraphics[scale=0.35]{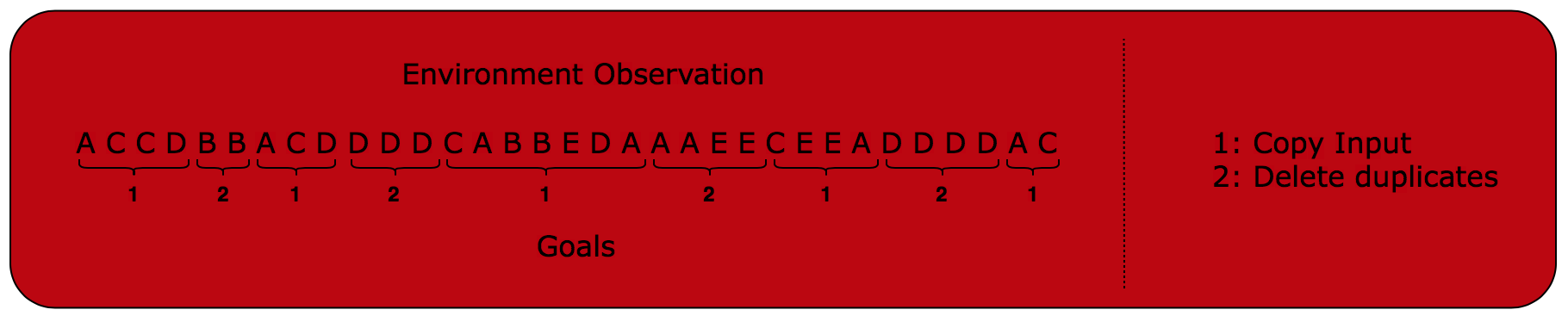}
  \caption{Demonstration of task, which consist of two subtasks. Model training to switch between polices, based on input, to reach the main goal.}
\end{figure*}%

\begin{table}
\centering
\begin{tabular}{lll}
\hline
                & Mean reached reward   & Number of steps \\ \hline
Copy            &         30            & 30000          \\  \hline
Drop duplicates &         16            & 55000           \\ \hline
\end{tabular}
\caption{The number of steps required to AFM to reach the target number of rewards in each of the subtasks}
\label{my-label}
\end{table}

\section{Settings}
For the classification task, the model had different size of forgetting layer in network $V$ and [784, 512, size of forgetting layer in $V$] for controller network $C$, where the Adam optimizer learning rate of 0.001 was used for both models $V$ and $C$. For an improved evolutionary algorithm, 100 populations and 100 generations were used, with a batch size of 64 for every network and 100 epochs for each sub-task training. Threshold drop hyperparameter for $E$ was set equal to $97\%$.

For the algorithmic task, the model had [27, 128, 20] neurons for forgetting network $V$, where hidden layers were forgetting layers, and [27, 128, 128] neurons for controller network $C$. Q learning with $\gamma$ = 0.99, experience replay size of 1000000, RMSprop optimizer with a 0.00025 learning rate, a 0.95 alpha, and a 0.01 eps were used for forgetting network $V$. Threshold target value for Copy task was set to mean 25 rewards in 100 trails, and 15 for Drop duplicates task. For controller $C$, binary cross entropy loss with an Adam optimizer learning rate set to 0.001 were used.

\section{Discussion}

Artificial intelligence systems are not yet able to cope with tasks with a deep hierarchy where people demonstrate quite acceptable results. This hypothesis demonstrates an alternative view of solving this problem. It is proposed that humans are able to solve deep hierarchy tasks using systems of active forgetting. If active forgetting systems are introduced into artificial intelligence systems, this hypothesis asserts the following:

\begin{itemize}
  \item Forgetting as universal hierarchical architecture.
  \item The paradox of planning uselessness.
\end{itemize}

\subsection{Universal hierarchical architecture}

\textbf{Hypothesis 1:} \textit{The processing of the environment hierarchy by the agent occurs due to the hierarchy of active neural forgetting processes. The group of neurons $F_{h}$ is allocated to the task, and a subgroup of these neurons, $F_{h_{h}}$,is allocated to the sub-task, where $h$ is a hierarchy level of tasks in the environment}

To solve the problem of tasks with a hierarchy, the existing AI architectures divide the tasks into levels [13]. With temporal abstraction, models are divided into two or more hierarchical stages, but it is impossible to clearly define a finite number of hierarchy levels. It is also not possible to ultimately determine exactly to which hierarchical level a particular task belongs.

With the AFM, operations will occur in the space of neurons. As was described, neurons are determined for each task, but while the model is learning for a new task, neurons are selected from the number of neurons that are used to solve a higher-level problem, This allows transfer learning of a new task more quickly because the neurons have been trained for a more general task in the same context.

\subsection{Planning exists only on paper} 

\begin{figure*}[t]
  \centering\includegraphics[scale=0.55]{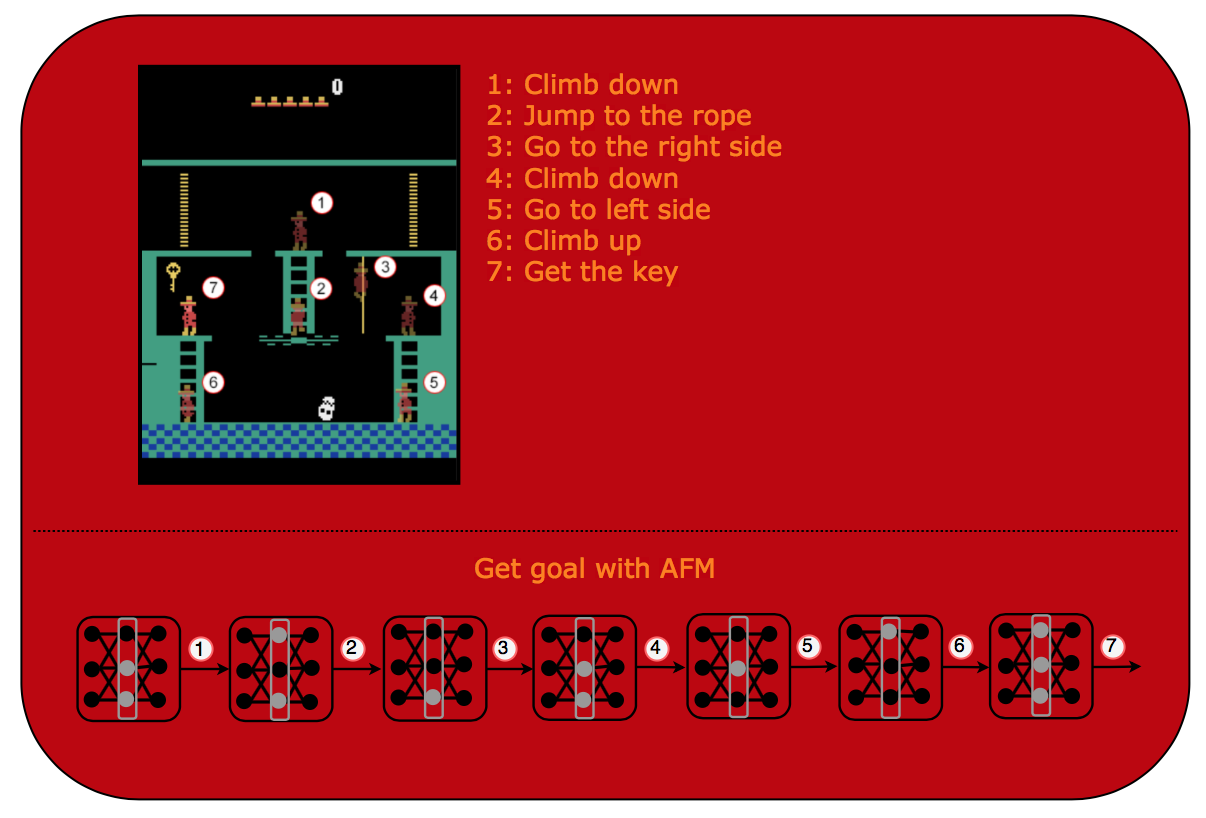}
  \caption{Illustration of MONTEZUMA’S REVENGE Atari 2600 solving, with Active Forgetting Machine model.}
\end{figure*}%

\textbf{Hypothesis 2:} \textit{If, based only on experience, the system is able to select a correct action in each situation, planning becomes unnecessary.}

In the classical concept, planning is the process of creating upcoming activities that allow for choosing the most correct actions. If the depth of the environment hierarchical partition is minimal, there is no problem perfectly planning which actions must be achieved to reach each goal. Since the hierarchical partition can be any depth, the sequence of predicted actions can be very long and expensive.

\textit{MONTEZUMA’S REVENGE} for the Atari 2600 is a suitable example for demonstrating the hypotheses. Using natural language guided reinforcement learning[5], this study presented implicit divisions of tasks with natural language instructions. If each set of instructions are presented as independent tasks, it is possible to move from the space of natural language to the space of neurons. Thereafter, when moving from sub-task to sub-task, certain neurons will be activated, which will lead the agent to the most beneficial activity in each of the sub-tasks, as shown in Figure 3. If the architecture is trained to perform each task qualitatively, it is clear that there is no need for planning within each sub-task. Based only on experience, the agent has the ability to perform actions, heading for the next goal. There is also no need to plan the sequence of sub-tasks that need to be performed because if the system achieves the goal in one sub-task, it means that it has passed correctly to the next sub-task, where the correctness of the sub-task has relevance to the main goal.

After all, it cannot be said that planning is not necessary at all. Reinforcement learning planning exists as an explicit activity that should be considered as one of the sub-tasks. In some cases, as in cases with other tasks, the agent fulfils the planning activity. Planning mechanisms are used to set goals, and may be used to qualitatively determine the group of neurons that need to be activated to solve the problem. Combining overcoming forgetting with Visual reinforcement Learning with Imagined Goals [12] can help models to set on some target and more correctly activate necessary group of neurons. This approach saves resources compared to planning, because all information about the necessary actions is stored in neurons.

\section*{References}
\medskip
\small
[1]  Kingma Diederik P, Salimans Tim, and Welling Max.  Variational dropout and the local reparameterization trick. \textit{arXiv preprint arXiv:1506.02557}, 2015.

[2]  H. Ebbinghaus. Memory:  A contribution to experimental psychology. \textit{New York by Teachers College}, 1913.

[3]  Chrisantha Fernando, Dylan Banarse, Charles Blundell, Yori Zwols, David Ha,  Andrei  A.  Rusu,  Alexander  Pritzel,  and  Daan  Wierstra.  
Pathnet: Evolution channels gradient descent in super neural networks. \textit{arXiv preprint arXiv: 1701.08734}, 2017.

[4]  Jiri Hron, Alexander G. de G. Matthews, and Zoubin Ghahramani.  Variational gaussian dropout is not bayesian. \textit{arXiv preprint arXiv: 1711.02989}, 2017.

[5]  Russell Kaplan, Christopher Sauer, and Alexander Sosa. Beating atari with natural language guided reinforcement learning. \textit{arXiv preprint arXiv: 1704.05539}, 2017.

[6]  Diederik  P  Kingma  and  Max  Welling.   Auto-encoding  variational  bayes. \textit{arXiv preprint arXiv: 1312.6114}, 2013.

[7]  James Kirkpatrick, Razvan Pascanu, Neil Rabinowitz, Joel Veness, Guillaume Desjardins, Andrei A. Rusu, Kieran Milan, John Quan, Tiago Ra-malho,  Agnieszka Grabska-Barwinska,  Demis Hassabis,  Claudia Clopath, Dharshan Kumaran, and Raia Hadsell. Overcoming catastrophic forgetting in neural networks. \textit{arXiv preprint arXiv: 1612.00796}, 2016.

[8]  Arun Mallya and Svetlana Lazebnik.  Packnet:  Adding multiple tasks to a single network by iterative pruning. \textit{arXiv preprint arXiv:1701.08734}, 2017.

[9]  Nicolas  Y.  Masse,  Gregory  D.  Grant,  and  David  J.  Freedman.   Alleviating  catastrophic  forgetting  using  context-dependent  gating  and  synaptic stabilization. \textit{arXiv preprint arXiv: 1802.01569v1}, 2018.

[10]  Volodymyr Mnih, Koray Kavukcuoglu, David Silver, Alex Graves, Ioannis Antonoglou,  Daan  Wierstra,  and  Martin  Riedmiller.   Playing  atari  with deep reinforcement learning. \textit{arXiv preprint arXiv: 1312.5602}, 2013.13

[11]  Dmitry  Molchanov,  Arsenii  Ashukha,  and  Dmitry  Vetrov.   Variational dropout sparsifies deep neural networks. \textit{arXiv preprint arXiv: 1701.05369}, 2017.

[12]  Ashvin Nair, Vitchyr Pong, Murtaza Dalal, Shikhar Bahl, Steven Lin, and Sergey Levine.  Visual reinforcement learning with imagined goals. \textit{arXiv preprint arXiv: 1807.04742},2018.

[13]  Sutton Richard S, Doina Precup, and Satinder Singh.  Between MDPs and semi-MDPs:  A framework for temporal abstraction in reinforcement learning. \textit{Artificial Intelligence 112:181–211 (1999)}, 1999.

[14]  Andrei A. Rusu, Neil C. Rabinowitz, Guillaume Desjardins, Hubert Soyer, James Kirkpatrick, Koray Kavukcuoglu, Razvan Pascanu, and Raia Hadsell.  Progressive neural networks. \textit{arXiv preprint arXiv: 1606.04671}, 2016.

[15]  Susan  Sangha,  Chloe  McComb,  and  Ken  Lukowiak.   Forgetting  and  the extension  of  memory  in  lymnaea. \textit{Journal  of  Experimental  Biology}  2003206:  71-77; doi:  10.1242/jeb.00061, 2003.

[16]  Joan  Serra,  Diıdac  Suris,  Marius  Miron,  and  Alexandros  Karatzoglou. Overcoming catastrophic forgetting with hard attention to the task. \textit{arXiv preprint arXiv: 1801.01423v2},2018.

\end{document}